\documentclass[11pt,a4paper]{article}

\usepackage{mathtools,cancel}

\usepackage{soul,xcolor}
\usepackage{xcolor}
\usepackage{multirow}
\usepackage{pstricks}
\usepackage{dcolumn}
\usepackage{bm}
\usepackage{latexsym}
\usepackage{dcolumn}
\usepackage[utf8x]{inputenc} 
\usepackage{amsmath}
\usepackage{amsfonts,amssymb}
\usepackage{graphicx,epsfig}
\usepackage{color}
\usepackage{psfrag}
\usepackage{amsthm}
\usepackage{ulem} 
\usepackage{soul} 
\usepackage{flushend}
\usepackage{titlesec}
\usepackage{slashed}
\usepackage{authblk}
\usepackage[font=footnotesize]{caption}
\usepackage[hmarginratio=1:1,top=32mm,columnsep=60pt]{geometry}

\newcommand{\bea}{\begin{eqnarray}}
\newcommand{\eea}{\end{eqnarray}}
\newcommand{\be}{\begin{equation}}
\newcommand{\ee}{\end{equation}}
\newcommand{\ba}{\begin{array}}
\newcommand{\ea}{\end{array}}

\def\beq{\begin{equation}}
\def\eeq{\end{equation}}
\def\bea{\begin{eqnarray}}
\def\eea{\end{eqnarray}}

\begin{document}

\unitlength = 1mm

\title{\textbf{Analyzing Fairness of Classification Machine Learning Model with Structured Dataset}}

\author{\normalsize Ahmed Rashed$^a$\footnote{amrashed@ship.ed}, Abdelkrim Kallich$^a$\footnote{ak2206@ship.edu}, and Mohamed Eltayeb$^{b,c}$\footnote{443059243@stu.iu.edu.sa  }}
\affil{\small
$^a$Department of Physics, Shippensburg University of Pennsylvania,\\
Franklin Science Center, 1871 Old Main Drive, Pennsylvania, 17257, USA\\
$^b$Islamic University of Madinah, Medina, Al Jamiah, Madinah 42351, Saudi Arabia\\
$^c$University of Khartoum, Khartoum, Al-Nil Avenue, Khartoum 11115, Sudan

 }
\date{}

{\let\newpage\relax\maketitle}

\begin{abstract}
\noindent \normalsize 
Machine learning (ML) algorithms have become integral to decision-making in various domains, including healthcare, finance, education, and law enforcement. However, concerns about fairness and bias in these systems pose significant ethical and social challenges. This study investigates the fairness of ML models applied to structured datasets in classification tasks, highlighting the potential for biased predictions to perpetuate systemic inequalities. A publicly available dataset from Kaggle was selected for analysis, offering a realistic scenario for evaluating fairness in machine learning workflows.

To assess and mitigate biases, three prominent fairness libraries-Fairlearn by Microsoft, AIF360 by IBM, and the What-If-Tool by Google-were employed. These libraries provide robust frameworks for analyzing fairness, offering tools to evaluate metrics, visualize results, and implement bias mitigation strategies. The research aims to assess the extent of bias in the ML models, compare the effectiveness of these libraries, and derive actionable insights for practitioners.

The findings reveal that each library has unique strengths and limitations in fairness evaluation and mitigation. By systematically comparing their capabilities, this study contributes to the growing field of ML fairness by providing practical guidance for integrating fairness tools into real-world applications. These insights are intended to support the development of more equitable machine learning systems.

\end{abstract}


\newpage

\section{Introduction}
\label{sec:introduction}
Machine learning algorithms are widely used in various domains, including entertainment, shopping, healthcare, finance, education, law enforcement, and high-stakes areas like loans [1] and hiring decisions [2, 3]. They provide advantages such as tireless performance and the ability to process numerous factors [4, 5]. However, algorithms can also exhibit biases, leading to unfair outcomes [6, 7]. Bias in machine learning can lead to discriminatory outcomes, especially when decisions directly affect individuals or communities. Addressing these issues is essential to ensure that machine learning systems operate ethically and equitably. Fairness in decision-making requires the absence of prejudice or favoritism based on inherent or acquired characteristics, and biased algorithms fail this standard by skewing decisions toward certain groups.

The concept of "fairness" in algorithmic systems is heavily influenced by the sociotechnical context. Various types of fairness-related harms have been identified:
\begin{enumerate}
\item	\textbf{Allocation Harm}: Unfair distribution of opportunities, resources, or information, such as an algorithm selecting men more often than women for job opportunities [8].
\item	\textbf{Quality-of-Service Harm}: Disproportionate failures affecting certain groups, e.g., facial recognition misclassifying Black women more often than White men [9], or speech recognition underperforming for users with speech disabilities [10].
\item	\textbf{Stereotyping Harm}: Reinforcement of societal stereotypes, such as image searches for "CEO" predominantly showing photos of White men [8].
\item	\textbf{Denigration Harm}: Offensive or derogatory outputs from systems, like misclassifying people as gorillas or chatbots using slurs [8].
\item	\textbf{Representation Harm}: Over- or under-representation of certain groups, e.g., racial bias in welfare fraud investigations or neglect of elderly populations in public-space monitoring [8].
\item	\textbf{Procedural Harm}: Decision-making practices violating social norms, such as penalizing job applicants for extensive experience or failing to provide transparency, justification, or appeals for algorithmic decisions [11].
\end{enumerate}
These harms often overlap and are not exhaustive, emphasizing the need for careful consideration of fairness from the development stage of algorithmic systems.

This research focuses on the fairness analysis of machine learning models when applied to structured datasets using classification problems. Structured datasets are widely utilized in machine learning applications due to their organized nature, which allows for efficient analysis and processing. However, biases in structured datasets, often stemming from historical inequalities or systemic discrimination, can propagate into ML models, necessitating robust fairness evaluation and mitigation strategies.

To conduct this study, a publicly available dataset [12] from Kaggle was selected. Kaggle datasets provide diverse and realistic scenarios for analyzing machine learning models, making them ideal for this type of research. The dataset was preprocessed and used to develop classification models, a common task in machine learning that involves predicting discrete labels based on input features. Classification problems are particularly relevant for fairness studies because biased predictions can disproportionately impact specific groups.

To evaluate and mitigate potential biases, three state-of-the-art fairness libraries were employed: Fairlearn by Microsoft, AIF360 by IBM, and the What-If Tool by Google. These libraries provide comprehensive toolsets for fairness analysis, including metrics to assess fairness, visualizations to interpret model behavior, and algorithms to mitigate bias. By leveraging these libraries, this research systematically evaluates the fairness of classification models and explores techniques to reduce bias in their predictions.

The objectives of this research are threefold:
\begin{enumerate}
\item	To assess the extent of bias present in machine learning models trained on the selected structured dataset.
\item	To compare the performance and effectiveness of the three fairness libraries in identifying and mitigating bias.
\item	To provide actionable insights into the application of fairness tools in real-world machine learning workflows.
\end{enumerate}
The remainder of this paper is organized as follows: Section 2 provides a review of related work on machine learning fairness. Section 3 describes the methodology, including the dataset, preprocessing steps, and model development process. Section 4 discusses the implementation of fairness analyses using the selected libraries and their capabilities. Section 5 presents a comparative analysis and results of the libraries. Finally, Section 6 concludes with a summary of findings.

\section{Review of Related Work}

Bias in machine learning (ML) models has been a growing area of concern, particularly as these models increasingly impact critical societal domains such as healthcare, hiring, and criminal justice. Numerous studies have explored the origins, manifestations, and mitigation strategies of bias, providing a comprehensive foundation for understanding and addressing this pervasive issue.

One key area of research focuses on identifying and characterizing biases in machine learning models. Ref. [13] provide a broad taxonomy of biases, categorizing them into historical, representation, and measurement biases. Historical bias originates from inequities in the data itself, even before ML techniques are applied. Representation bias emerges when certain groups are under- or over-represented in the training data, leading to skewed model predictions [14]. Measurement bias arises when the features or labels used for training do not accurately reflect the target variable due to flawed measurement processes.

Another stream of work has delved into bias detection methods. Techniques such as disparate impact analysis [15] and fairness metrics like demographic parity, equal opportunity, and disparate mistreatment [16] have become standard tools. For structured datasets, researchers often focus on quantifying group fairness and individual fairness. Group fairness ensures equitable treatment across predefined demographic groups, while individual fairness emphasizes treating similar individuals similarly [17]. Ref. [18] discusses how these fairness definitions often conflict, necessitating trade-offs tailored to specific applications.

The literature also emphasizes the technical challenges of mitigating bias. One popular approach involves pre-processing techniques to address biases within the dataset itself. For example, Ref. [19] proposes re-weighting data samples or modifying labels to ensure fairness before model training. In-processing methods, such as adversarial debiasing [20], introduce fairness constraints directly into the training process. Post-processing techniques adjust the model outputs to achieve fairness metrics, such as the re-ranking methods proposed by [16]. However, Ref. [21] highlights the inherent trade-offs between fairness metrics, illustrating that achieving fairness often requires sacrificing accuracy.

Bias analysis in structured datasets, specifically, has garnered attention due to the widespread use of tabular data in decision-making systems. Structured datasets often carry latent biases stemming from historical inequities in human decision-making or systemic discrimination. The COMPAS dataset, used in criminal justice, exemplifies these challenges, with studies showing racial disparities in predictive outcomes [22]. Research on structured datasets also highlights the role of feature selection and data preprocessing in amplifying or mitigating biases. Ref. [23] examines how feature correlation with sensitive attributes impacts fairness, proposing strategies for disentangling these relationships.

Recent work has explored interpretability and its role in bias analysis. Ref. [24] introduced LIME (Local Interpretable Model-agnostic Explanations) to help stakeholders understand model predictions, aiding the detection of biased decision-making patterns. In Ref. [25] further developed SHAP (SHapley Additive exPlanations), which provides consistent and locally accurate feature importance values. These tools have been instrumental in identifying bias within structured datasets, as they enable granular analyses of how individual features contribute to unfair predictions.

Additionally, researchers are increasingly incorporating intersectionality into bias studies. Ref. [26] emphasized the importance of evaluating models across multiple demographic axes, demonstrating how performance disparities can compound for intersectional groups, such as Black women in facial recognition systems. For structured datasets, studies by [27] propose fairness-enhancing interventions that consider multiple subgroups simultaneously, avoiding the pitfalls of single-axis fairness analysis.

The literature on bias in machine learning models spans a wide range of topics, from foundational definitions and detection methods to mitigation strategies and interpretability tools. While significant progress has been made, challenges remain in applying these techniques to structured datasets, particularly in balancing fairness with other competing objectives such as accuracy and interpretability. This review underscores the importance of continued research into holistic and context-sensitive approaches for analyzing and mitigating bias in machine learning models.

\section{Dataset and Model Details}
\label{sec:meth}
The study utilized the Adult Income dataset from the UCI Machine Learning Repository [12], which is publicly available and commonly used for classification tasks. The dataset contains demographic and income-related attributes, aiming to predict whether an individual’s income exceeds \$50,000 annually. Key features include age, education, occupation, work hours, marital status, race, and gender, alongside a binary target variable indicating income class.

To process the dataset, categorical variables were encoded using techniques such as one-hot encoding, and continuous variables were normalized to ensure compatibility with the machine learning models. The dataset was split into training and testing subsets using an 80-20 split to evaluate model performance.

For the model, the pipeline employed several machine learning algorithms, including Logistic Regression, Decision Trees, and Gradient Boosting. Gradient Boosting, specifically implemented using XGBoost, emerged as the most effective algorithm for the classification task. Hyperparameter tuning was performed using grid search to optimize the model’s performance.

The study evaluated the model using standard classification metrics such as accuracy, precision, recall, and F1-score. In addition, fairness metrics such as demographic parity and disparate impact were calculated to analyze potential biases. Results highlighted disparities in prediction accuracy across demographic groups, with notable differences between male and female individuals and among racial groups. These findings underscored the importance of incorporating fairness evaluations into traditional performance assessments for machine learning models.

\section{Implementation of Fairness Analyses }

In recent years, the increasing reliance on machine learning (ML) in various sectors has led to growing concerns over fairness and bias in classification models. As these models can significantly influence decision-making processes in critical areas such as healthcare, finance, and criminal justice [28], ensuring their fairness has become imperative. Bias in ML models can manifest due to various factors, including skewed training data, model selection, and underlying societal biases, ultimately leading to discriminatory outcomes against marginalized groups [29].

To address these challenges, several libraries and tools have been developed to assist practitioners in analyzing and mitigating bias in their models. Among them, Fairlearn, AIF360, and What-If Tool stand out as comprehensive resources that offer unique functionalities for fairness evaluation and enhancement.

\begin{itemize}
\item \textbf{Fairlearn} [30]: Developed by Microsoft, this toolkit helps data scientists assess and improve the fairness of their AI models by providing a suite of metrics to evaluate fairness and algorithms for mitigating unfairness. Fairlearn emphasizes the importance of both social and technical dimensions of fairness in AI systems. It facilitates the understanding of how different aspects of a model contribute to disparities among groups.
\item \textbf{AIF360} [31]: This comprehensive library offers a wide range of metrics for assessing fairness and techniques for mitigating bias across the entire AI application lifecycle. Developed by IBM, AIF360 includes methods that can be integrated into different stages of the machine learning pipeline to facilitate fairness-aware modeling. It includes metrics for evaluating fairness across different societal demographics and offers re-parameterization strategies to improve model robustness.
\item \textbf{What-If-Tool} [32]: Created by Google, this interactive visualization tool allows users to explore and analyze machine learning models without requiring any coding. It supports performance testing in hypothetical scenarios, facilitating the understanding and explanation of model behavior. It enables users to observe model performance across different demographics and explore various "what-if" scenarios. It supports users in understanding how changes in input features affect model predictions, thus empowering them to conduct deeper bias analysis.
\end{itemize}

Together, these tools are essential for researchers aiming to understand and mitigate bias in machine learning classification models, equipping them with the methodologies to ensure equitable AI systems and informed decision-making processes.

\section{Results and Discussions}

In this section, we discuss the results and outcome of each fairness library. The main objective of our analysis is to reduce/improve the fairness metric and maintain or improve the performance metric. The codes and details can be found in [33]. The sensitive feature that we measured the bias in is gender. We investigate the model if it is biased against any group (male/female) in gender.

\textbf{For Fairlearn}, we used accuracy to measure the model's overall performance in predicting income classification and demographic parity difference as a fairness metric to evaluate the bias in the machine learning model. Accuracy provided a clear assessment of the predictive capability, while the demographic parity difference quantified disparities in outcomes across different demographic groups.

To address and mitigate the detected biases, we applied mitigation algorithms at three stages of the machine learning pipeline: preprocessing, in-processing, and postprocessing. Preprocessing techniques involved modifying the training data to reduce bias before feeding it into the model. For example, sensitive features in a dataset may be correlated with non-sensitive features. The Correlation Remover addresses this by eliminating these correlations, while preserving as much of the original data as possible, as evaluated by the least-squares error. In-processing methods integrated fairness constraints into the model training process, with approaches such as exponentiated gradient ensuring that the model learned fairer decision boundaries. Postprocessing focused on adjusting the predictions after model training, ensuring that the final outputs adhered to fairness criteria without retraining the model such as threshold optimizer which is built to satisfy the specified fairness criteria exactly and with no remaining disparity [34, 35, 36].

Beyond these individual techniques, we also explored the use of combined mitigation approaches, where two algorithms were applied in series to enhance fairness outcomes. For instance, preprocessing adjustments were complemented by postprocessing tweaks, leading to improved alignment with both accuracy and fairness objectives. This combined approach aimed to leverage the strengths of each mitigation stage to produce more equitable and reliable model predictions.

The evaluation process identified the best algorithm as the one that achieved a dual objective: maximizing performance metrics such as accuracy, precision, and recall while minimizing bias metrics like demographic parity difference. This comprehensive evaluation ensured that the selected method not only improved predictive power but also promoted equitable treatment across demographic groups. Our findings highlight the importance of an integrated approach to fairness, where multiple strategies are utilized in conjunction to address complex biases inherent in structured datasets.

Table 1 summarizes our results. The baseline model, prior to implementing any mitigation algorithm, achieved a performance metric (Accuracy) of 0.85 and a fairness metric (Demographic Parity Difference) of 0.30. Ideally, the accuracy would be 1.00, and the demographic parity difference would be 0.00. To address bias, we applied mitigation algorithms both individually and sequentially, using two different algorithms at distinct stages of the machine learning lifecycle. The best results were obtained using the exponential gradient algorithm, which maintained an accuracy of 0.85-only a 1.10\% decline compared to the baseline-and reduced the demographic parity difference to 0.05, an 83\% improvement over the baseline. Notably, similar results were achieved when combining the exponential gradient algorithm with the threshold optimizer.\\

\begin{center}
\includegraphics[scale=0.7]{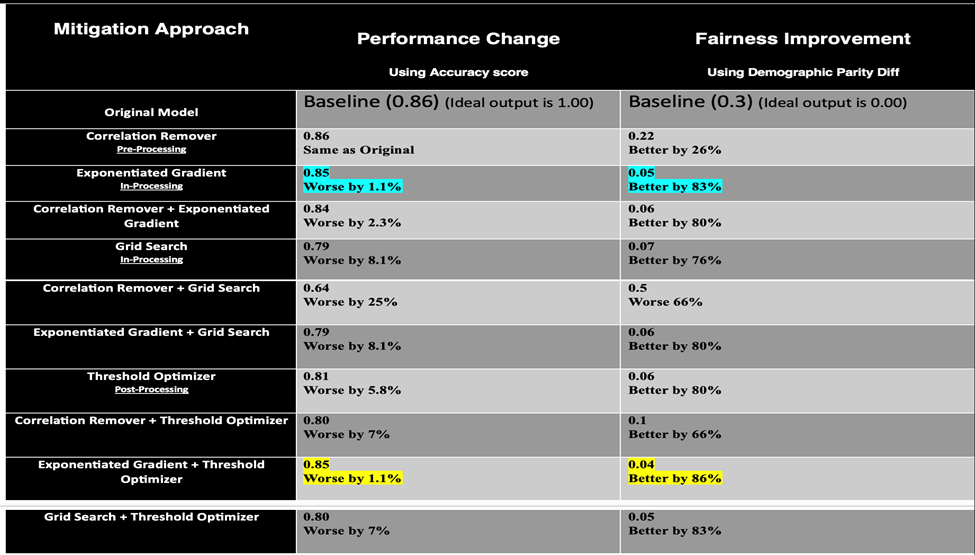}
Table 1. The results by applying Fairlearn library to the classification model. 
\end{center}

\textbf{For AIF360}, we employed two distinct mitigation algorithms within each stage of the machine learning pipeline to comprehensively address bias. In the preprocessing stage, Reweighing was identified as the most effective technique, where it assigned weights to instances based on their representation in different demographic groups. This approach ensured a balanced distribution of data, directly addressing biases embedded in the training dataset.

In the post-processing stage, Equalized Odds proved to be particularly impactful. This algorithm-imposed constraints during model training to ensure that predictive outcomes were not disproportionately distributed across sensitive attributes such as race or gender. By enforcing parity in true positive and false positive rates, Equalized Odds enhanced fairness without significantly compromising model accuracy.

We selected the Average Odds Difference (AOD) as the model performance metric, which was measured at 0.09 for the baseline model, compared to the ideal value of 0.00. For the fairness metric, we used the Statistical Parity Difference (SPD), which had a baseline value of 0.13, whereas the ideal value is also 0.00.

The applications of Reweighing and Equalized Odds yielded superior results, as these methods effectively reduced both bias and prediction error. By complementing the adjustments made during data preprocessing with fairness constraints integrated into the model training process, this approach addressed multiple facets of bias simultaneously. This synergy between algorithms ensured that the final model exhibited not only improved fairness metrics but also maintained robust performance on traditional classification metrics. For the Reweighing algorithm, AOD is 0.007 which is 8.3\% better than the baseline value and SPD is 0.01 with 12\% better than the baseline value. For the Equalized Odds algorithm, AOD is 0.0001 which is 9\% better than the baseline value and SPD is 0.0 with 13\% better than the baseline value. See Table 2 for the results.

Our findings underscore the importance of selecting appropriate mitigation strategies tailored to specific stages of the machine learning pipeline. By leveraging the strengths of Reweighing and Equalized Odds, we demonstrated that it is possible to achieve a balanced trade-off between fairness and accuracy, highlighting the potential of integrated approaches to bias mitigation in structured datasets.\\
 
\begin{center}
\includegraphics[scale=0.7]{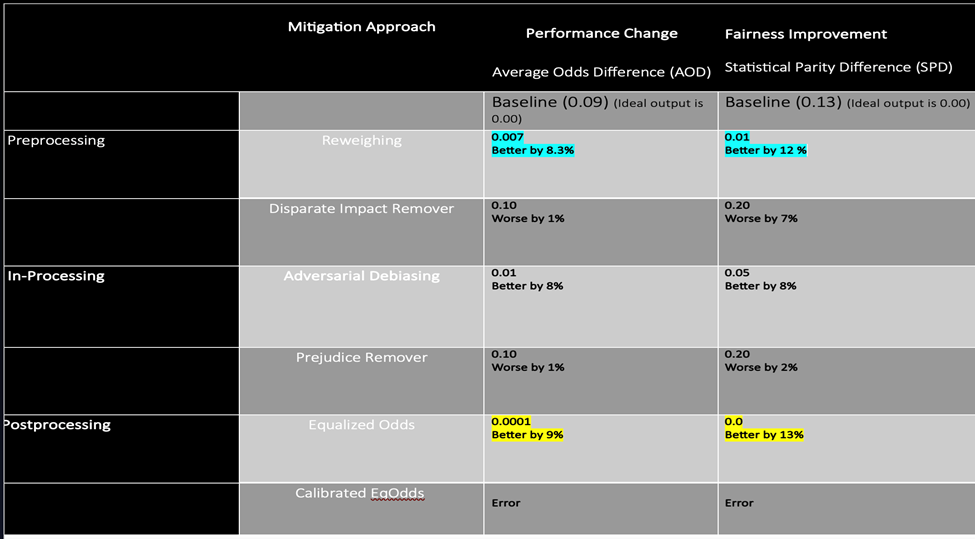} 
Table 2. The results by applying AIF360 library to the classification model. 
\end{center}

\textbf{In the What-If Tool}, adjusting the threshold for the labeled class impacts both the model's performance and its bias metrics. The optimal thresholds identified for this model were 0.2 and 0.4. At these thresholds, the model performance metric (Accuracy) increased from 0.33 to an average of 0.62, representing a 29\% improvement, while the bias metric (Demographic Parity Difference) decreased from 0.19 to 0.01, reflecting an 18\% reduction. These results are summarized in Table 3.\\

\begin{center}
\includegraphics[scale=0.7]{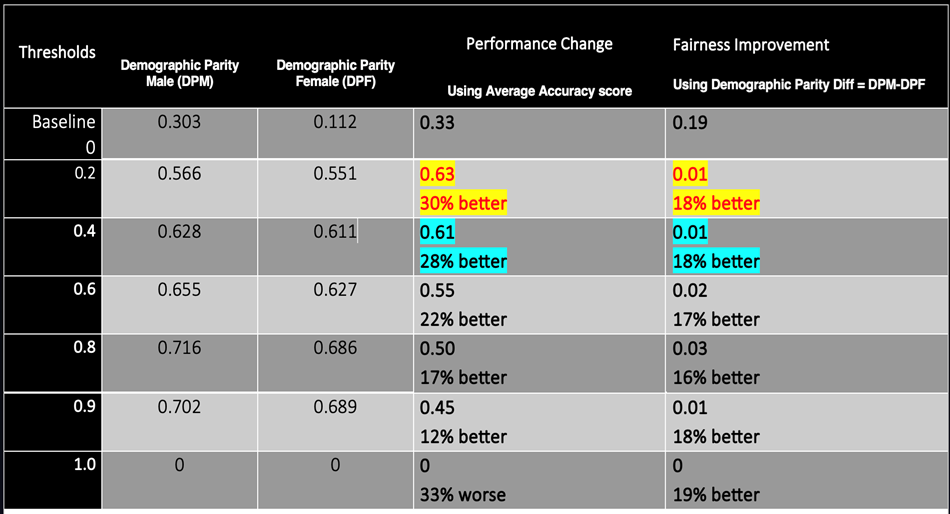}
Table 3. The results by applying What-If-Tool library to the classification model. 
\end{center}

\section{Conclusion}

This study examined the fairness of machine learning models in classification tasks using structured datasets, focusing on how biased predictions can reinforce systemic inequalities. A Kaggle dataset was analyzed to provide a realistic scenario, utilizing three fairness libraries—Fairlearn (Microsoft), AIF360 (IBM), and the What-If Tool (Google)—to evaluate and mitigate bias. The research compared these libraries' effectiveness, highlighting their unique strengths and limitations. The findings offered practical guidance for integrating fairness tools into machine learning workflows, contributing to the development of more equitable AI systems.

\section*{Acknowledgement}

We would like to thank Yale Center for Research Computing for supporting A.K. through the CAREERS project administered by the PSU Institute for Computational and Data Sciences (ICDS) under the National Science Foundation Award with No. 2018873.

\section*{REFERENCES}
\noindent
[1] Amitabha Mukerjee, Rita Biswas, Kalyanmoy Deb, and Amrit P. Mathur. 2002. Multi–objective evolutionary algorithms for the risk–return trade–off in bank loan management. Int. Trans. Oper. Res. 9, 5 (2002), 583–597.

\noindent
[2] Miranda Bogen and Aaron Rieke. 2018. HelpWanted: An Examination of Hiring Algorithms, Equity and Bias. Technical Report. Upturn.

\noindent
[3] Lee Cohen, Zachary C. Lipton, and Yishay Mansour. 2019. Efficient candidate screening under multiple tests and implications for fairness. arXiv:cs.LG/1905.11361 (2019).

\noindent
[4] Shai Danziger, Jonathan Levav, and Liora Avnaim-Pesso. 2011. Extraneous factors in judicial decisions. Proc. Nat. Acad. Sci. 108, 17 (2011), 6889–6892.

\noindent
[5] Anne O’Keeffe and Michael McCarthy. 2010. The Routledge Handbook of Corpus Linguistics. Routledge.

\noindent
[6] Julia Angwin, Jeff Larson, Surya Mattu, and Lauren Kirchner. 2019. Machine bias: There’s software used across the country to predict future criminals. and it’s biased against blacks. https://www.propublica.org/article/machine-biasrisk-assessments-in-criminal-sentencing.

\noindent
[7] Cathy O’Neil. 2016.Weapons of Math Destruction: How Big Data Increases Inequality and Threatens Democracy. Crown Publishing Group, New York, NY.

\noindent
[8] M. A. Madaio, L. Stark, J. Wortman Vaughan, and H. Wallach. Co-Designing Checklists to Understand Organizational Challenges and Opportunities around Fairness in AI. Chi 2020, pages 1–14, 2020.

\noindent
[9] J. Buolamwini and T. Gebru. Gender shades: Intersectional accuracy disparities in commercial gender classification. In Conference on fairness, accountability and transparency, pages 77–91, 2018.

\noindent
[10] A. Guo, E. Kamar, J. W. Vaughan, H. M. Wallach, and M. R. Morris. Toward fairness in AI for people with disabilities: A research roadmap. CoRR, abs/1907.02227, 2019. URL http://arxiv.org/abs/1907.02227.

\noindent
[11] C. Rudin, C. Wang, and B. Coker. The age of secrecy and unfairness in recidivism prediction. pages 1–46, 2018. URL http://arxiv.org/abs/1811.00731.

\noindent
[12] Dataset used in the study https://www.kaggle.com/code/jieyima/income-classification-model 

\noindent
[13] Mehrabi, N., Morstatter, F., Saxena, N., Lerman, K., \& Galstyan, A. (2021). A Survey on Bias and Fairness in Machine Learning. ACM Computing Surveys, 54(6), 1-35.

\noindent
[14] Suresh, H., \& Guttag, J. V. (2021). A Framework for Understanding Unintended Consequences of Machine Learning. Communications of the ACM, 64(8), 62-71.

\noindent
[15] Feldman, M., Friedler, S. A., Moeller, J., Scheidegger, C., \& Venkatasubramanian, S. (2015). Certifying and Removing Disparate Impact. Proceedings of the 21st ACM SIGKDD International Conference on Knowledge Discovery and Data Mining (KDD).

\noindent
[16] Hardt, M., Price, E., \& Srebro, N. (2016). Equality of Opportunity in Supervised Learning. Proceedings of the 30th Conference on Neural Information Processing Systems (NeurIPS).

\noindent
[17] Dwork, C., Hardt, M., Pitassi, T., Reingold, O., \& Zemel, R. (2012). Fairness through Awareness. Proceedings of the 3rd Innovations in Theoretical Computer Science Conference.

\noindent
[18] Binns, R. (2018). Fairness in Machine Learning: Lessons from Political Philosophy. Proceedings of the 2018 Conference on Fairness, Accountability, and Transparency (FAT).

\noindent
[19] Kamiran, F., \& Calders, T. (2012). Data Preprocessing Techniques for Classification without Discrimination. Knowledge and Information Systems, 33(1), 1-33.

\noindent
[20] Zhang, B. H., Lemoine, B., \& Mitchell, M. (2018). Mitigating Unwanted Biases with Adversarial Learning. Proceedings of the 2018 AAAI/ACM Conference on AI, Ethics, and Society (AIES).

\noindent
[21] Chouldechova, A. (2017). Fair Prediction with Disparate Impact: A Study of Bias in Recidivism Prediction Instruments. Big Data, 5(2), 153-163.

\noindent
[22] Angwin, J., Larson, J., Mattu, S., \& Kirchner, L. (2016). Machine Bias. ProPublica.

\noindent
[23] Xu, D., Yuan, S., Zhang, L., \& Wu, X. (2020). FairGAN: Fairness-aware Generative Adversarial Networks. Proceedings of the 2020 International Joint Conference on Artificial Intelligence (IJCAI).

\noindent
[24] Ribeiro, M. T., Singh, S., \& Guestrin, C. (2016). "Why Should I Trust You?" Explaining the Predictions of Any Classifier. Proceedings of the 22nd ACM SIGKDD International Conference on Knowledge Discovery and Data Mining (KDD).

\noindent
[25] Lundberg, S. M., \& Lee, S.-I. (2017). A Unified Approach to Interpreting Model Predictions. Proceedings of the 31st Conference on Neural Information Processing Systems (NeurIPS).

\noindent
[26] Buolamwini, J., \& Gebru, T. (2018). Gender Shades: Intersectional Accuracy Disparities in Commercial Gender Classification. Proceedings of the 2018 Conference on Fairness, Accountability, and Transparency (FAT).

\noindent
[27] Kearns, M., Neel, S., Roth, A., \& Wu, Z. S. (2018). Preventing Fairness Gerrymandering: Auditing and Learning for Subgroup Fairness. Proceedings of the 35th International Conference on Machine Learning (ICML).

\noindent
[28] Barocas, S., Hardt, M., \& Narayanan, A. (2019). Fairness and Accountability in Machine Learning. Proceedings of the 2019 Conference on Fairness, Accountability, and Transparency.

\noindent
[29] Angwin, J., Larson, J., Mattu, S., \& Kirchner, L. (2016). Machine Bias: There's Software Used Across the Country to Predict Future Criminals. And it's Biased Against Blacks. ProPublica.

\noindent
[30] Fairlearn by Microsoft. (n.d.). Retrieved from https://fairlearn.org/ 

\noindent
[31] AIF360 by IBM. (n.d.). Retrieved from https://aif360.mybluemix.net/ 

\noindent
[32] What-If Tool by Google. (n.d.). Retrieved from https://github.com/google/tf-what-if 

\noindent
[33] Github of the project: https://github.com/mohammad2012191/Fairness-in-Machine-Learning-Identifying-and-Mitigation-of-Bias/ 

\noindent
[34] Moritz Hardt, Eric Price, and Nati Srebro. Equality of opportunity in supervised learning. In NeurIPS, 3315–3323. 2016. \\
URL: https://proceedings.neurips.cc/paper/2016/hash/9d2682367c3935defcb1f9e247a97c0d-Abstract.html 

\noindent
[35] Hilde Weerts, Lambèr Royakkers, and Mykola Pechenizkiy. Does the end justify the means? on the moral justification of fairness-aware machine learning. arXiv preprint arXiv:2202.08536, 2022.

\noindent
[36] Brent Mittelstadt, Sandra Wachter, and Chris Russell. The unfairness of fair machine learning: levelling down and strict egalitarianism by default. arXiv preprint arXiv:2302.02404, 2023.


\end{document}